%% file: eventbert.tex
\documentclass[11pt,a4paper]{article}
\usepackage[hyperref]{acl2021}
\usepackage{times}
\usepackage{graphicx}
\usepackage{tabu}
\usepackage{booktabs}
\usepackage{latexsym}

\usepackage{algorithm}
\usepackage{algorithmic}
\usepackage{booktabs}       
\usepackage{amsfonts}       
\usepackage{nicefrac}       
\usepackage{microtype}      
\usepackage{xcolor}         
\usepackage{subfigure}
\usepackage{mathtools}
\usepackage{amsmath}
\usepackage[algo2e]{algorithm2e} 
\usepackage{wrapfig}
\usepackage{graphbox}

\newcommand{\dataset}{{\usefont{T1}{pzc}{m}{n} ART}}

\input{math_commands}

\DeclareMathOperator*{\transformerenc}{Transformer-Enc}
\DeclareMathOperator*{\pool}{Pool}
\DeclareMathOperator*{\mlp}{MLP}
\DeclareMathOperator*{\mysigmoid}{Sigmoid}

\aclfinalcopy 
\def\aclpaperid{3668} 

\title{EventBERT: A Pre-Trained Model for Event Correlation Reasoning}

\author{Yucheng Zhou$^{1}$\thanks{~~Work is done during internship at Microsoft.} ,
        Xiubo Geng$^{2\dagger}$, Tao Shen$^{2}$, Guodong Long$^{3}$, Daxin Jiang$^{2}$\thanks{~~Corresponding author.} \\
         $^{1}$Fudan University, Shanghai, China \\
         $^{2}$Microsoft, Beijing, China \\
         $^{3}$Australian AI Institute, School of CS, FEIT, University of Technology Sydney \\
         {\tt yczhou18@fudan.edu.cn, \{xigeng,shentao,djiang\}@microsoft.com}\\
         {\tt guodong.long@uts.edu.au}
         }

\date{}

\begin{document}
\maketitle

\begin{abstract}
Event correlation reasoning infers whether a natural language paragraph containing multiple events conforms to human common sense. For example, ``\textit{Andrew was very drowsy, so he took a long nap, and now he is very alert}'' is sound and reasonable. In contrast, ``\textit{Andrew was very drowsy, so he stayed up a long time, now he is very alert}'' does not comply with human common sense. Such reasoning capability is essential for many downstream tasks, such as script reasoning, abductive reasoning, narrative incoherence, story cloze test, etc. However, conducting event correlation reasoning is challenging due to a lack of large amounts of diverse event-based knowledge and difficulty in capturing correlation among multiple events. In this paper, we propose EventBERT, a pre-trained model to encapsulate eventuality knowledge from unlabeled text. Specifically, we collect a large volume of training examples by identifying natural language paragraphs that describe multiple correlated events and further extracting event spans in an unsupervised manner. We then propose three novel event- and correlation-based learning objectives to pre-train an event correlation model on our created training corpus. Empirical results show EventBERT outperforms strong baselines on four downstream tasks, and achieves SoTA results on most of them. Besides, it outperforms existing pre-trained models by a large margin, e.g., 6.5$\sim$23\%, in zero-shot learning of these tasks.
\end{abstract}

\section{Introduction}
\begin{figure}[t]
    \centering
    \includegraphics[width=0.95\linewidth]{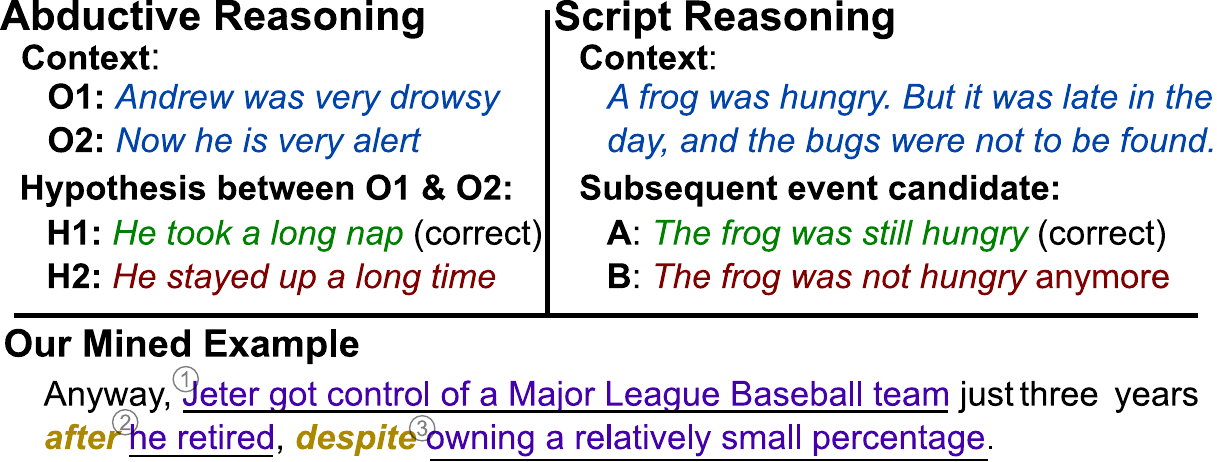}
    \caption{\small Examples of two downstream event-related reasoning tasks, i.e., abductive reasoning (left) and script reasoning (right), as well as our created corpus (bottom). }
    \label{fig:intro_case}
\end{figure}

Event correlation reasoning is critical for AI systems since it benefits many downstream tasks. Figure~\ref{fig:intro_case} lists several examples of event-related reasoning tasks. Figure~\ref{fig:intro_case}(left) is an example of abductive reasoning, which aims to infer the most plausible explanation for incomplete observations. Given the two observations, ``\textit{Andrew was very drowsy}'' and ``\textit{Now he is very alert}'', we could infer that ``\textit{he took a long nap}'' is more plausible than ``\textit{he stayed up a long time}''. Another typical task in Figure~\ref{fig:intro_case}(right) is script reasoning. Given a sequence of events, ``\textit{A frog was hungry. But it was late in the day, and the bugs were not to be found}'', we need to infer a potential subsequent event -- ``\textit{The frog was still hungry}''. 
Although formulated as different tasks, they all require event correlation\footnote{In this paper, ``correlation'' refers to associated or discourse relations (including causal, temporal) among events if not specified.} reasoning, i.e., to infer whether a paragraph containing multiple events conforms to human common sense. Here \textit{event} refers to a span involving a predicate with its arguments and describing either states of entities/things or how they act in real world, e.g., ``\textit{A frog was hungry}'', ``\textit{Bugs was not to be found}'', and ``\textit{The frog was still hungry}'' in Figure~\ref{fig:intro_case}(right).

A major challenge for event correlation reasoning is that they require large amounts of diverse commonsense knowledge describing correlation among events. Although such information may exist in large-scale unlabeled corpus, from which modern pre-trained language models (i.e., BERT \citep{Devlin2019BERT}, RoBERTa \citep{Liu2019RoBERTa}, SpanBERT \citep{Joshi2020SpanBERT}) are learned, the text data is often noisy and thus cannot be used directly. 
First, events in the same paragraph may not necessarily have strong correlation. For example, although there are multiple events in the paragraph ``\textit{She was referring to several large cellular towers at one corner of the Sheriff's complex. Beside the towers was a large, power generator—or, at least, what we assumed to be the power generator.}'' (from book, \textit{Days Alone}),
there is no strong logical correlation among them.
Moreover, there is no explicit labeling for event boundary. Therefore, traditional masked language models (MLM) or span-based models are not aware of events. Another paradigm is to leverage human-curated paragraph with related events. \citet{sap2019atomic} collected 9 types of if-then event-based commonsense knowledge by crowd-sourcing. But, this paradigm is not scalable enough in both diversity and corpus size due to cost limitation.


In this paper, we propose EventBERT, a pre-trained model for general event correlation commonsense reasoning from unlabeled text. 
We propose a novel approach to automatically collect a large event correlation corpus.
Specifically, we first identify natural language paragraphs which contain multiple events with strong correlation with each other. We then explicitly mark event spans in those paragraphs. Figure~\ref{fig:intro_case}(bottom) shows an example paragraph with three identified events (i.e., ``\textit{Jeter got control of a Major League Baseball team}'', ``\textit{he retired}'', and ``\textit{owning a relatively small percentage}''). Furthermore, we present three novel self-supervised contrastive learning objectives, i.e., correlation-based event ranking, contradiction event tagging and discourse relation ranking, to learn correlations among events effectively.

EventBERT has 3 advantages: 1) Since the training corpus stems from a large corpus of unlabeled data, it provides large amounts of diverse event correlation knowledge, and can be applied to a variety of downstream tasks. 2) The recognition of events in paragraphs enables the pre-trained model to be aware of the events and further understand the correlation among them better. 
3) The event-based self-supervised learning objectives enforce the model to focus on event correlation instead of token-level concurrence as in MLM.

We evaluate EventBERT on multiple downstream tasks, including script reasoning, abductive commonsense reasoning, narrative incoherence detection, and story cloze test. While only continually pre-trained on mined \textsc{BookCorpus} (that has been used in previous pre-training) with very few number of updates (180k steps within 500 GPU hours), it outperforms strong baselines on 4 benchmark datasets, and achieves state-of-the-art (SoTA) results on most of them. EventBERT outperforms existing pre-trained models by 6.5\%\textasciitilde23\% in a zero-shot setting. Moreover, with only a small size (e.g., 10\% for story cloze test) of task-specific training data, it can achieve similar performance of the baseline using all data. All these results demonstrate that EventBERT encapsulates rich eventuality knowledge for event-correlation reasoning.


\section{Related Work}

\paragraph{Masked Language Modeling. }
Most recent works \citep{Devlin2019BERT,Liu2019RoBERTa,He2020DeBERTa} pre-train an encoder via self-supervised masked language modeling (MLM) \citep{Devlin2019BERT}
but sole word-level masking makes the model focus on local context.
Hence, this masking scheme is extended to other text units beyond words, e.g., phrase \citep{Sun2019ERNIE1}, named entity \citep{Sun2019ERNIE1} and random span \citep{Joshi2020SpanBERT}. 
It is intuitive to further extend into event-level one to capture event correlation, e.g., CoCoLM \citep{Yu20CoCoLM}, but there are two major obstacles: 
First, training objectives with MLM still focus more on token-level concurrence. For example, given a context [\textit{Andrew was very drowsy, so ?, and now he is very alert}], both candidates ``\textit{he took a long nap}'' and ``\textit{he stayed up a long time}'' are assigned with similar perplexities by following \citet{Davison2019Mining}. 
Second, an event is expressed with a sentence or a clause, so masking such an extremely long span hinders MLMs from learning the context from both sides. 

\paragraph{Sentence-level Pre-training. } 
Several objectives learn inter-sentence semantics in text, 
e.g., next sentence prediction (NSP) \citep{Devlin2019BERT}, sentence-order prediction (SOP) \citep{Lan2020AlBERT}, 
sentence permutation \citep{Lewis2020BART}. Although NSP and SOP share a very high-level inspiration with our work in terms of training example composition, they are unlikely to explicitly acquire event correlations. 
First, event correlations scarcely exist in a sentence from open-domain corpora, not to mention unconsciousness of event boundary. 
Second, they operate at sentence level, but ignore that a discourse relation (e.g., \textit{but}, \textit{then}), coupled with event(s) to compose a sentence and serving as indispensable component of cross-event correlation, should be considered individually. 
Third, negatives are sampled from corpus wide (NSP) and intra-document (SOP), so trivial and less diverse. 

\paragraph{Commonsense-centric Pre-training.} 
Some works learn concept-centric commonsense knowledge. 
One paradigm is designing specific continual pre-training objectives. 
\citet{Shen2020GLM} leverage ConceptNet \citep{Speer2017ConceptNet} to guide masking spans for MLM.
Similar to denoising in BART, \citet{Zhou21Pre} merely consider token-level concepts (e.g., verb and noun) and aims to recover original sentence given a set of concepts or disordered sentence. 
Another paradigm is transfer learning from similar commonsense tasks, such as Unicorn \citep{Lourie21UNICORN} and UnifiedQA \citep{Khashabi20UnifiedQA}, which need extra human-labeled data. 

\paragraph{Event-based Pre-training.}
Similarly, some approaches inject pre-trained models with rich event-related knowledge, which focus more on event correlations than commonsense-centric ones. 
First, some approaches like COMET \citep{Bosselut2019COMET,hwang2020atomic2020} conduct continual pre-training on triple facts from a human-curated knowledge graph (KG) to build neural knowledge models. 
The KG is usually related to eventuality, e.g., ATOMIC \citep{sap2019atomic}, 
which depends on crowd-sourcing and thus limits the model's scalability. 
Moreover, the knowledge model, learned on the triple fact (i.e., a pair of events with a prompt), focuses merely on pairwise correlation, thus degrading the general-purpose pre-trained model. 
Second, some other approaches mine event information in raw corpus instead of the KGs.  
DEER \citep{Han2020DEER} performs continual pre-training via temporal and event masking predictions, i.e., new masking schemes for MLM, to focus on event temporal relations. 
\citet{Lin2020Conditional} leverage BART structure and propose to recover a temporally-disordered or event-missing sequence to the original one by a denoising autoencoder. 
\citet{Wang20CLEVE} use AMR structure to design self-supervised objectives but focus only on event detection task. 
In contrast, we focus on long-term event correlations underlying raw corpus and targets correlation-based reasoning tasks. 




\section{EventBERT} \label{sec:meth}

\begin{figure}[t]
    \centering
    \includegraphics[width=0.45\textwidth]{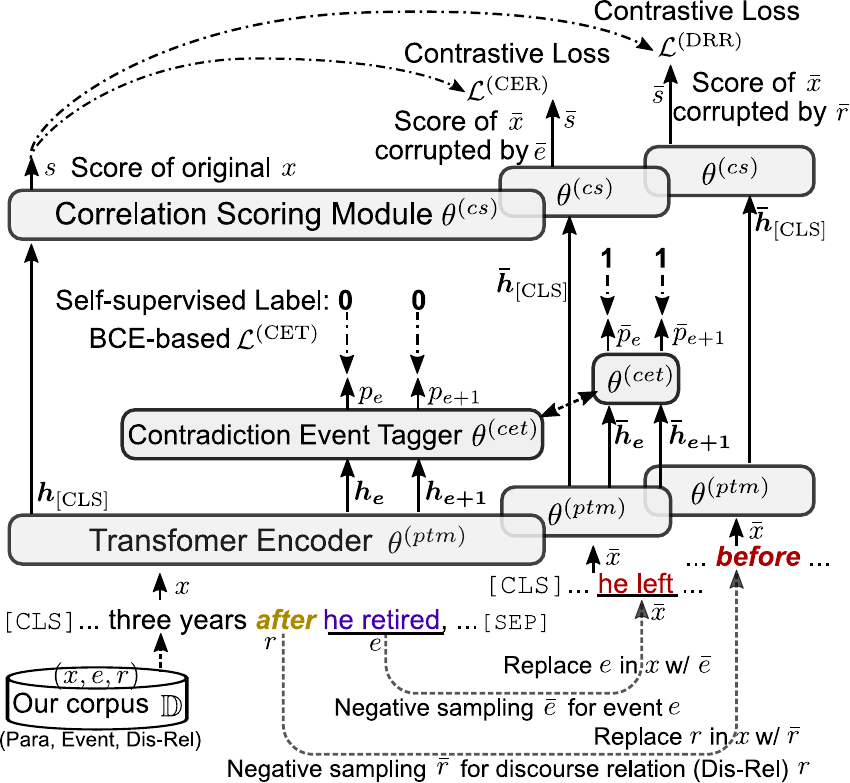}
    \caption{\small An overview of our proposed EventBERT with three self-supervised contrastive learning objectives.}
    \label{fig:model_illustration}
\end{figure}

This section begins with data collection (\S\ref{sec:meth-train_data_collect}), followed by self-supervised objectives for training (\S\ref{sec:meth-eventbert_pretrain} and Figure~\ref{fig:model_illustration}).
Lastly, we detail two knowledge transfer paradigms (\S\ref{sec:meth-eventbert_eval}).

\subsection{Training Data Collection} \label{sec:meth-train_data_collect}


\paragraph{Text Corpus and Filtering.} 
We leverage \textsc{BookCorpus} \citep{Zhu2015bookcorpus}, which contains 16 different sub-genres books (e.g., romance, historical), as training corpus. 
Compared to other text corpora stemming from encyclopedia (e.g., \textsc{Wikipedia}), news \citep{Nagel2016Ccnews} and web crawl \citep{Gokaslan2019Openwebtext}, \textsc{BookCorpus} is a large collection of books (11K books with 74M sentences with 1B words) and likely to contain diverse event correlation and rich eventuality knowledge. 
Then, we conduct corpus filtering to ensure our training data rich in event correlation. 
The filtering unit is set to ``paragraph'' since it usually contains a complete storyline or reasoning logic for learning efficacy whilst consists of mere several sentences to keep filtering efficiency. Besides basic filtering strategies like removing unreadable and meaningless text pieces, we propose to filter paragraphs and keep those with strong event correlation according to discourse relation keywords (e.g., \textit{however}, \textit{while}). Specifically, we filter out paragraphs without any of the pre-defined connectives in PDTB\footnote{\url{https://www.seas.upenn.edu/~pdtb/}}, based on our observation that paragraphs with these connectives are prone to contain strong correlation among events and vice versa. Considering that an event is usually centered with a verb (e.g., ``\textit{Andrew was very drowsy}'', ``\textit{he stayed up a long time}''), we further restrict that these keywords should be adjacent to a verb in the dependency parsing tree to avoid false positive cases where these keywords does not express connectives in some paragraphs. We detail our discourse-verb association filtering strategy in Algorithm~\ref{alg:paragraph_filter}. 
After filtering, we get 199.9M (out of 1B) words in 3.9M paragraphs. 

\begin{algorithm}[!t] \small
    \SetKwInOut{Input}{Input}
    \SetKwInOut{Output}{Output}
    \Input{~\textsc{BookCorpus}, Discourse Relations $\sK$ from PDTB.}
    \Output{~Filtered paragraphs, $\sF$.}
    $\sF \leftarrow \{\}$; \tcp{To store filtered paragraphs}
    \For{Para $\in$ \textsc{BookCorpus} \tcp{Traverse paragraph}}{
      $\sM \leftarrow \{\}$; \tcp{To store meta information}
      $\gG \leftarrow \text{Dep-Parsing}(\textit{Para})$ \par
      $\sV \leftarrow \text{Verbs in \textit{Para} according to PoS-Tagging}(\textit{Para})$ \par
      $\sL \leftarrow$ Locate all discourse relations in \textit{Para} through lexicon-based matching with $\sK$;\par
      \For{$r\in\sL$ \tcp{Traverse discourse rel in \textit{Para}}}{
        $\sN\leftarrow~\text{Neighbor words of}~r~\text{over}~\gG$; \par
        \For{$v\in \sN \!\cap\! \sV$ \tcp{Discourse-verb connection}}{
          $\sM \leftarrow \sM \cup \{(r, v)\}$
        }
      }
      \lIf{$|\sM| > 0$}{
        $\sF \leftarrow \sF \cup \{(\textit{Para}, \sM)\}$
      }
    }
    \caption{\small Discourse-verb association filtering} \label{alg:paragraph_filter}
\end{algorithm}

\paragraph{Event Extraction and Training Set Building.}
Then, it is essential to extract events in text since explicit event annotations will facilitate designs of unsupervised learning objectives towards event correlation reasoning. 
Particularly, event extraction aims to find an event trigger (e.g., a predicate) and then retrieve all arguments (e.g., subject, adverb, preposition) for the trigger. 
Previous works usually employ supervised event extraction models \citep{Wadden2019EventExtract} and semantic role labeling \citep{Huang2018SRL4EventEx} but both suffer from high overheads. 
Thus, for high efficiency, we focus only on verb-triggered event and resort to low-level syntactic features of a sentence, says dependency parsing tree \citep{Chen2014DepParse} that describes modification relations between words. 
Coupled with Part-of-Speech (PoS) tagging, we can readily extract events from a sentence: 
We first take the word with PoS of ``\textit{Verb}'' as an event trigger and then retrieve a sub-tree, whose root is the trigger, from the dependency parsing tree.
Considering the dependency tree is ordered, the retrieved verb-rooted sub-tree can be mapped to a span of words and regarded as an extracted event. 
We finally generate training examples for EventBERT as detailed in Algorithm~\ref{alg:train_set_gen}. 
Consequently, each training example $(x, e, r)$ denotes a paragraph, $x$, containing an event $e$ with a discourse relation $r$ ($r\in x \wedge r\notin e$). 
It is worth mentioning (1) $e$ is a word span in $x$ so $x$ can also be written as $x=[x^{(fw)}, e, x^{(bw)}]$, and (2) $r$ is usually a single word while sometimes consecutive words. 

\begin{algorithm}[!t] \small
    \SetKwInOut{Input}{Input}
    \SetKwInOut{Output}{Output}
    \Input{~Filtered corpus $\sF$.}
    \Output{~EventBERT Training Set $\sD$.}
    $\sD \leftarrow \{\}$; \tcp{To store training examples} 
    \For{(Paragraph, $\sM$) $\in \sF$}{
      $\gG \leftarrow \text{Dep-Parsing}(\textit{Paragraph})$ \par
      \For{$(r, v) \in \sM$ \tcp{discourse relation and verb}} 
      {
        $\gG' \leftarrow$~$v$-rooted sub-tree from $\gG$; \tcp{extract event}
        $e \!\leftarrow\!$~Map~$\gG'$ to span in \textit{Paragraph}; \!\!\tcp{\!\!\!extract event}
        $x \leftarrow \textit{Paragraph}$ \tcp{Define context}
        $\sD \leftarrow \sD\cup \{(x,  e, r)\}$;
      }
    }
    \caption{\small Training set building with event extraction}\label{alg:train_set_gen}
\end{algorithm}

\subsection{Self-supervised Learning Objectives}\label{sec:meth-eventbert_pretrain}

In line with most pre-trained models \citep{Devlin2019BERT}, we use the Transformer encoder \citep{Vaswani2017Transformer} to produce contextualized embeddings. 
Instead of learning from scratch, we adopt MLM-based pre-trained Transformer encoder (e.g., BERT \citep{Devlin2019BERT}, RoBERTa \citep{Liu2019RoBERTa}) and focus on further injecting event correlation knowledge. 
For this purpose, we propose three self-supervised contrastive learning objectives, i.e., \textit{Correlation-based Event Ranking}, \textit{Contradiction Event Tagging} and \textit{Discourse Relation Ranking}.
The former two teach the model to distinguish the correct event against negative ones based on event correlations within paragraphs, while the latter one helps the model identify subtle difference among discourse relations. 
But it is still non-trivial to generate challenging, diverse negative events. 

\paragraph{Event-based Negative Sampling.} Instead of negative sampling by a generative model like GPT \citep{Radford2019GPT2}, we retrieve diverse events from $\sD$ to avoid pattern gap between human-written and machine-generated texts. Thereby, we build an event pool $\sQ=\{e|\forall (x,e,r)\in\sD\}$ and propose three heuristic schemes to derive candidates of negative events for $e$ (e.g., $e=$``\textit{he looks very worried}'') in $(x,e,r)$: 
\begin{enumerate} 
    \item \textit{Lexicon-based Retrieval}: We retrieve events from $\sQ$ according to lexicon overlap with $e$, leading to nuanced negative events $\sQ^{(lb)}$ (e.g., ``\textit{he dies}'') and thus challenging distractors. 
    \item \textit{PoS-based Retrieval}: We retrieve $\bar e$ from $\sQ$ according to overlaps of PoS tags between $e$ and $\bar e$, leading to syntax-similar negative events $\sQ^{(pb)}$ (e.g., ``\textit{it tastes pretty good}''). 
    \item \textit{In-domain Retrieval}: We retrieve neighbor events around $e$ up to five paragraphs, leading to in-domain negative events $\sQ^{(id)}$ (e.g., ``\textit{he moves to my side}'').
\end{enumerate}
We remove $e$ itself from every $\sQ^{(*)}$ and limit the size of $\sQ^{(*)}$ to a $N$ ($N=3$). 
Then, for the positive event $e$. we sample a negative one $\bar e$ from $\sQ^{(lb)}$, $\sQ^{(pb)}$ and $\sQ^{(id)}$ with probabilities of $20\%$, $60\%$ and $20\%$ respectively. 
A behind intuition is, lexicon-based retrieval may returns false negative samples whilst in-domain retrieval probably returns trivial negative samples. 
Lastly, we extract $M$ negative samples $\{(\bar x, \bar e, r)\}_{k=1}^M$ for $(x, e, r)$, where $\bar x\coloneqq [x^{(fw)}, \bar e, x^{(bw)}]$ and $M=5$ in our experiments. 

\paragraph{Correlation-based Event Ranking. } 
Based on our training example $(x,e,r)$ and corresponding negative samples $\{(\bar x, \bar e, r)\}_{k=1}^M$, we propose the first self-supervised contrastive learning objective, named correlation-based event ranking (CER), to rank the correlation-correct paragraph $x$ against the paragraph $\bar x$ corrupted by a distracting event $\bar e$. 
Formally, we first pass $x=[w_1,\dots,w_n]$ and $\bar x=[\bar w_1,\dots,\bar w_m]$ into the Transformer encoder to generate contextualized embeddings, i.e., 
\begin{align}
    &\mH = [\vh_1, \dots, \vh_n]
    \coloneqq \transformerenc( x; \theta^{(ptm)}), \label{eq:transformer_enc_pos} \\
    &\notag \bar\mH^{(k)} = [\bar\vh^{(k)}_1, \dots, \bar\vh^{(k)}_m] \label{eq:transformer_enc_neg} \\
    &~~~~~~~~\notag  \coloneqq \transformerenc(\bar x^{(k)}; \theta^{(ptm)}), \\
    &~~~~~~~~~~\forall k\in[1,M], 
\end{align}
where $\mH\in\R^{d\times n}$, $\bar\mH^{(k)}\in\R^{d\times m}$, superscript $k$ in parenthesis indicates the index of negative samples, and $x$ ($\bar x^{(k)}$ also) has been prefixed and suffixed with special tokens,
\texttt{[CLS]} and \texttt{[SEP]}, respectively. Note the above two Transformer encoders are parameter-tied and parameterized by $\theta^{(ptm)}$. Then, we pool a sequence of word-level contextualized embeddings (i.e., $\mH$ or $\bar\mH^{(k)}$), to sequence-level representation:
\begin{align}
    \notag  \vv \coloneqq& \vh_{\texttt{[CLS]}} = \pool(\mH)\in\R^{d},~~\text{and}\\
    \bar\vv^{(k)} \coloneqq& \bar\vh^{(k)}_{\texttt{[CLS]}} = \pool(\bar\mH^{(k)})\in\R^{d},~\forall k\in[1,M], \label{eq:transformer_pooling}
\end{align}
where, $\pool(\cdot)$, as defined by \citet{Devlin2019BERT}, denotes collecting the embedding of \texttt{[CLS]} to represent the entire sequence. 
Next, the sequence-level representation is passed into a one-way multi-layer perceptron (MLP) to derive a correlation score of the targeted event in a paragraph, 
\begin{align}
    s =& \mlp(\vv; \theta^{(cs)})~~\text{and}~~\bar s^{(k)}=\mlp(\bar\vv^{(k)}; \theta^{(cs)}), \label{eq:correlation_score}
\end{align}
where $\forall k\in[1,M]$, $s$ and $\bar s^{(k)}\in\R$, and $\mlp(\cdot; \theta^{(cs)})$ denotes our correlation scoring module. 
Lastly, to facilitate transfer to downstream multi-choice question answering, we regard this ranking problem as a contrastive classifying objective and define a negative log likelihood loss, i.e., 
\begin{align}
    \notag  \gL^{(\text{CER})} =& - \sum\nolimits_\sD \log\!P^{(cer)}(x |x, \{\bar x^{(k)}\}_{k=1}^M) \\
    =& \softmax([s; \bar s^{(1)}; \dots; \bar s^{(M)}])_1 \label{eq:loss_cer}
\end{align}  
where subscript ``$1$'' indicates the first dim of $\softmax$ distribution, corresponding to the positive $x$.

\paragraph{Contradiction Event Tagging. } 
Moreover, we take a step closer to the corrupted paragraphs and determine whether a word belongs to a negative-sampled event or the original one.
Therefore, we define another self-supervised learning objective based on contrastive examples, called contradiction event tagging (CET), to build a binary classifier at word level. 
To be formal, we first adopt the sequences of contextualized embeddings from Eq.(\ref{eq:transformer_enc_pos}-\ref{eq:transformer_enc_neg}), i.e., $\mH= [\vh_1, \dots, \vh_n]$ and $\bar\mH^{(k)}=[\bar\vh^{(k)}_1, \dots, \bar\vh^{(k)}_m]$. 
And then, we employ another one-way MLP to generate a probability, which measures if a word is contradictory to contexts in the paragraph, based on each word-level contextualized embedding, i.e., 
\begin{align}
    P^{(cet)}\left(\vh\right) =& \mysigmoid(\mlp(\vh ; \theta^{(cet)})),
\end{align}
$\forall\vh\in \{\mH\}\cup\{\bar\mH^{(k)}\}_{k=1}^M$. Finally, we can define a binary cross-entropy loss for contradiction event tagging objective:
\begin{align}
    \notag  \gL^{(\text{CET})} &= - \sum_{(x,e,r)\in\sD} \biggl( 
    \sum_{i\in e}\log \left(1-P^{(cet)}\left(\vh_i\right)\right) \\
    &+ \sum_{\{(\bar x, \bar e, r)\}_{k=1}^M}\sum_{i\in\bar e}\log P^{(cet)}\left(\bar\vh^{(k)}_i\right) \biggr). \label{eq:loss_cet}
\end{align}

\paragraph{Discourse Relation Ranking. } 

To further exploit event-correlation information underlying paragraphs, it is promising to consider another kind of negative sampling from the perspective of ``discourse relation''. 
That is, we can sample a couple of negative discourse relations $\bar r$ to corrupt $r$ in the paragraph $x$, and then employ our model to distinguish them. 
To this end, we propose the third self-supervised contrastive learning objective, discourse relation ranking (DRR), with a very similar target with CER -- ranking the original paragraph over the corrupted ones. 
Thereby, we share learnable parameters in CER with those in DRR here to improve correlation-based ranking and enhance the correlation scoring module. 
Specifically, we first sample $M$ negative relations $\bar r$ from PDTB keywords $\sK$ for $(x,e,r)$. To avoid false negative, we remove the keywords with the same categories\footnote{A taxonomy of discourse relations (e.g., `\textit{after}' for temporal, `\textit{so}' for causal) is defined in Appendix~C of PDTB manual.} as $r$ from $\sK$. 
Also, we denote the $\bar r$-corrupted paragraph as $\bar x$, where the corruption is achieved by \textit{word} (sometimes \textit{phrase}) replacement. 
Then, we pass $x$ and $\bar x$ individually into the Transformer encoder $\theta^{(ptm)}$ as in Eq.(\ref{eq:transformer_enc_pos}-\ref{eq:transformer_pooling}), followed by the correlation scoring module $\theta^{(ce)}$ as in Eq.(\ref{eq:correlation_score}). 
Lastly, identical to Eq.(\ref{eq:loss_cer}) in CER, we can define DRR's training loss as
\begin{align}
    \gL^{(\text{DRR})} = - \sum\nolimits_\sD \log P^{(drr)}(x |x, \{\bar x^{(k)}\}_{k=1}^M). \label{eq:loss_drr}
\end{align}

Consequently, the learning objective of EventBERT is to minimize
$\gL = \gL^{(\text{CER})}+\gL^{(\text{CET})}+\gL^{(\text{DRR})}$.
\subsection{Knowledge Transfer to Downstream Task} \label{sec:meth-eventbert_eval}
After pre-trained, EventBERT can be readily transferred to a wide range of event-centric downstream tasks. 
Since EventBERT is equipped with rich event-correlation knowledge by contrastive learning, the transfer can be achieved by either zero-shot evaluation or supervised fine-tuning.

\paragraph{Zero-shot Evaluation. } 
It is vitally significant to conduct zero-shot evaluations of a pre-trained model on downstream tasks since this can verify if the targeted knowledge was successfully injected into the pre-trained model. 
Empowered by the well-trained correlation scoring module $\theta^{(cs)}$, it is straightforward to adapt EventBERT into a zero-shot scenario for various-formatted tasks because $\theta^{(cs)}$ coupled with $\theta^{(ptm)}$ can assign a text with a score to measure if the text correctly expresses event correlations. 
Take multi-choice or Cloze-type question answering as an example. It aims at choosing an answer $a^{*}$ from a candidate set $\sA$ to best fit into a given context $[x^{(fw)}, ?,  x^{(bw)}]$ -- assigning the highest score to $[x^{(fw)}, a^{*}, x^{(bw)}]$. 
In this case, EventBERT can directly derive plausible scores for all candidate answers and then return the answer with highest score, i.e., 
\begin{align}\label{eq:event_zero_shot}
    \notag a^* = \arg&\max\nolimits_{a\in\sA} \mlp\Bigl(\pool\bigl(\transformerenc\\
    &~~~~~~([x^{(fw)}, a, x^{(bw)}];\theta^{(ptm)})\bigr);\theta^{(ce)}\Bigr). 
\end{align}
In contrast, traditional MLM-based pre-training cannot be directly applied to such a task, but by following \citet{Davison2019Mining}, we can adopt an autoregression-like operation to greedily generate the words masked by a candidate answer, where ``autoregression-like'' means we only recover a single \texttt{[mask]} back to the original word in each feed-forward process and then repeat, and ``\textit{greedily}'' means the recovered word is assigned with highest MLM probability in each process. 
Thus, an MLM-based pre-trained model can perform zero-shot evaluation on this task by
\begin{align}\label{eq:roberta_zero_shot}
    \notag a^* = \arg\max\nolimits_{a\in\sA} P^{(mlm)}&(a~~|~~[x^{(fw)}, \texttt{[MASK]},\\
    & \dots, \texttt{[MASK]},  x^{(bw)}]),
\end{align}
where the number of \texttt{[MASK]} is equal to word number in the corresponding $a$.


\paragraph{Supervised Fine-tuning. } Following the common practice in most pre-trained Transformers \citep{Devlin2019BERT,Liu2019RoBERTa}, we can fine-tune $\theta^{(ptm)}$ from EventBERT with a task-specific prediction module $\theta^{(task)}$ on a task in a supervised manner. For example, given a multi-choice question answering task again, we pass each candidate $a\in\sA$ with the context $[x^{(fw)}, ?,  x^{(bw)}]$ into $\pool(\transformerenc(\cdot;\theta^{(ptm)}))$ to produce a sequence-level representation, and then feed the representation into an MLP-based scorer, $\mlp(\cdot;\theta^{(task)})$, for a plausible score of the candidate. Here, $\theta^{(task)}$ is random initialized and then updated with $\theta^{(ptm)}$ towards the task-specific learning objective during fine-tuning.


\begin{table}[t]\small
\centering
\resizebox{\columnwidth}{!}{
\begin{tabular}{lc}
\toprule
\textbf{Method}                 & \textbf{ACC (\%)} \\ \midrule
SGNN + Int\&Senti \citep{Ding2019Event}              & 56.03             \\
\midrule
Random                          & 20.00             \\
PMI \citep{Chambers2008Unsupervised}                          & 30.52             \\
Event-Comp \citep{Wilding2016What}                     & 49.57             \\
SGNN \citep{li2018constructing}                           & 52.45             \\
RoBERTa-base \citep{Liu2019RoBERTa}                   & 56.23             \\
RoBERTa-large \citep{Liu2019RoBERTa}                  & 61.53             \\ 
RoBERTa + knwl. \citep{Lv2020Integrating}                 & 58.66             \\
EventBERT (\textbf{ours})                      & \textbf{63.50}    \\ \bottomrule
\end{tabular}}
\caption{\small Fine-tuning on script reasoning. First part involves extra human-craft data, e.g., curated KG or transfer from other tasks. }
\label{tab:sr}
\end{table}

\section{Experiments}

\paragraph{Datasets.}
We evaluate on 4 datasets for 4 downstream tasks, i.e., MCNC \citep{li2018constructing} for script reasoning, \emph{\dataset{}} \citep{bhagavatula2019abductive} for abductive commonsense reasoning, ROCStories \citep{mori2020finding} for narrative incoherence detection, and story cloze test \citep{mostafazadeh2016corpus}. All these are independent of our pre-trained corpus.
\begin{itemize}
    \item \textbf{Multi-choice narrative cloze (MCNC).} Given an event chain, it aims to select the most plausible subsequent event from 5 candidates. We follow the official data split
    with 140,331/10,000/10,000 samples in training/dev/test sets. 
    \item \textbf{\emph{\dataset{}}.} We evaluate the abductive NLI task on the \emph{\dataset{}} dataset. Given two observations about the world, it aims to select the most plausible explanatory hypothesis from two choices. We also follow the official data split
    with 17,801/1,532/3,059 examples for training/dev/test.
    \item \textbf{ROCStories.} ROCStories is a well-organized story-generation task. We follow \citep{mori2020finding} to use it for narrative incoherence detection, where one random sentence is removed for each five-sentence story. 
    The goal is to predict the missing position. We use the same data split
    as \citet{mori2020finding} where there are 78,528/9,816/9,817 examples in the training/dev/test sets.
    \item \textbf{Story Cloze Test.} Based on ROCStories. 
    it aims to choose the right ending from two alternative endings for a four-sentence context. We follow the official data split by \citet{mostafazadeh2016corpus}, and there are 98,161/1,871/1,871 examples in the training/dev/test sets.
\end{itemize}

\begin{table}[t]\small
\centering
\begin{tabular}{lc}
\toprule
\textbf{Method}             & \textbf{ACC (\%)} \\ \midrule
McQueen* \citep{Mitra2019Exploring}     & 84.18             \\
UNICORN \citep{Lourie21UNICORN}     & 78.30             \\
\midrule
Random*                     & 50.41             \\
BERT-base* \citep{Devlin2019BERT}                  & 63.62             \\
BERT-large* \citep{Devlin2019BERT}                 & 66.75             \\
RoBERTa-large* \citep{Liu2019RoBERTa}              & 83.91             \\
HighOrderGN* & 82.04             \\
CALM \citep{Zhou21Pre}     & 77.12             \\
EventBERT (\textbf{ours})                   & \textbf{84.51}             \\ \bottomrule
\end{tabular}
\caption{\small Fine-tuning results on abductive commonsense reasoning.  *from the leaderboard. First part involves extra human-craft data.}
\label{tab:acr}
\end{table}


\begin{table}[t]
\centering
\resizebox{0.95\columnwidth}{!}{
\begin{tabular}{lc}
\toprule
\textbf{Method}  & \textbf{ACC (\%)} \\ \midrule
Random           & 20.00             \\
RoBERTa-large \citep{Liu2019RoBERTa}   & 73.94             \\ 
Max-pool Context \citep{Yusuke2020Finding} & 35.00             \\
GRU Context \citep{Yusuke2020Finding}      & 52.20             \\ 
EventBERT (\textbf{ours})        & \textbf{75.03}    \\ \bottomrule
\end{tabular}}
\captionof{table}{\small Fine-tuning results on narrative incoherence detection. }
\label{tab:nid}
\end{table}

\begin{table}[t]
\centering
\resizebox{\columnwidth}{!}{
\begin{tabular}{lc}
\toprule
\textbf{Method}          & \textbf{ACC (\%)} \\ \midrule
Random                   & 50.00             \\
HCM \citep{chaturvedi2017story}  & 77.60             \\
val-LS-skip \citep{srinivasan2018simple}             & 76.50             \\
Finetuned Transformer LM \citep{radford2018improving} & 86.50             \\
RoBERTa-large \citep{Liu2019RoBERTa}           & 87.10             \\ 
EventBERT (\textbf{ours})               & \textbf{91.33}    \\ \bottomrule
\end{tabular}}
\captionof{table}{\small Fine-tuning on story cloze test.}
\label{tab:sct}
\end{table}

\begin{figure}[t]
\begin{center}
\centering
\begin{minipage}[t]{\linewidth}
\centering
\includegraphics[width=0.47\linewidth]{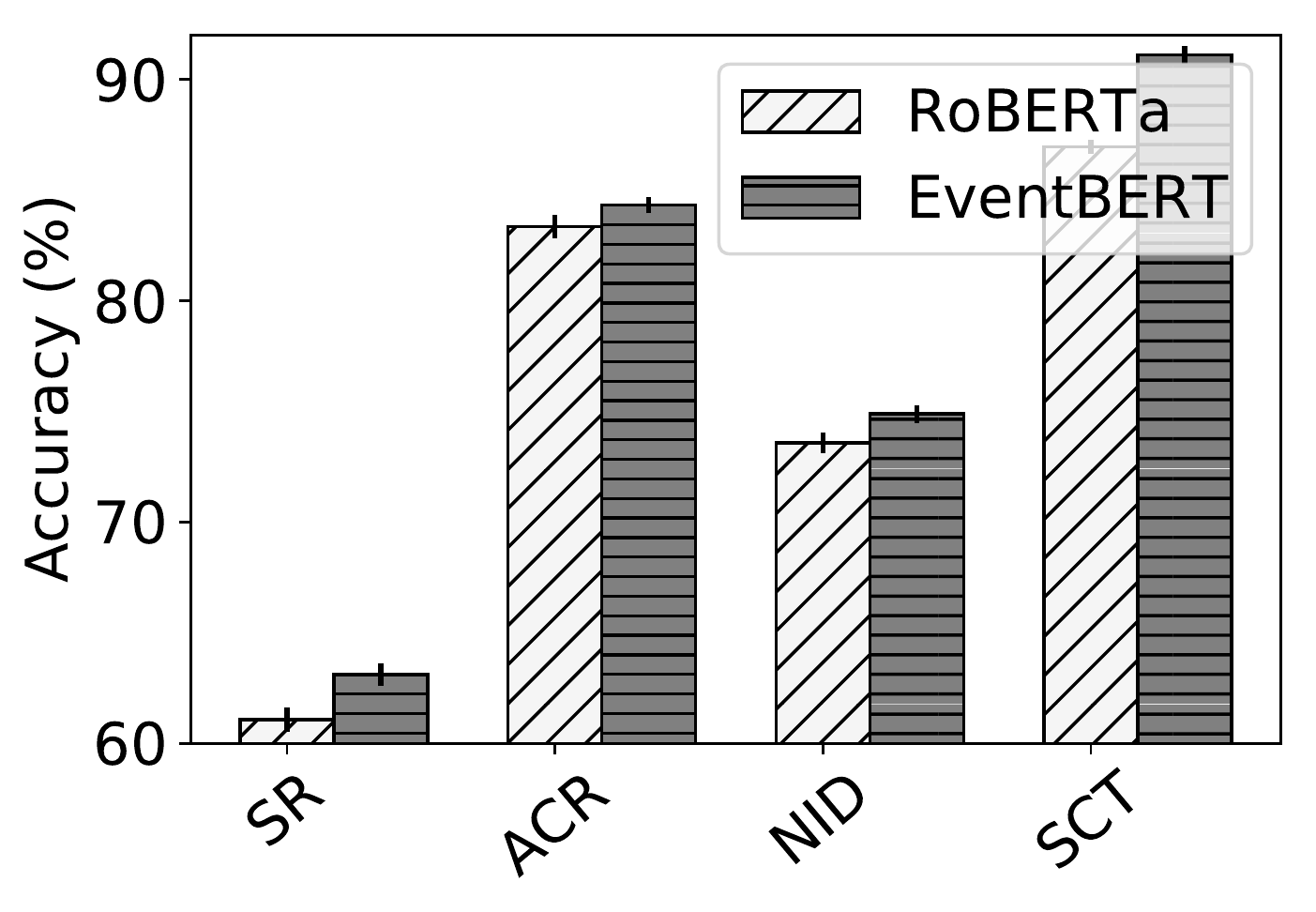}
\includegraphics[width=0.51\linewidth]{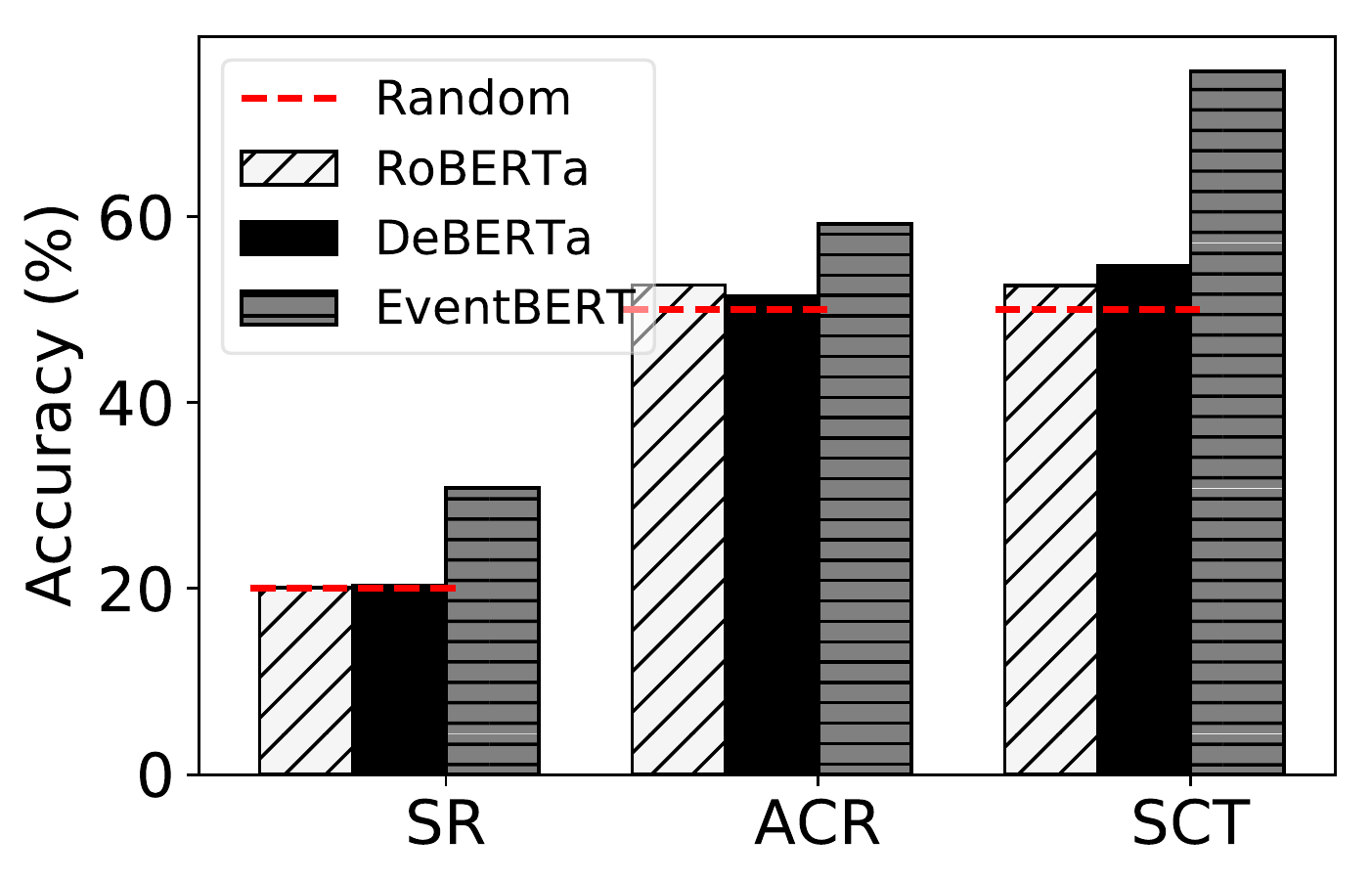}
\caption{\small \textbf{Left}: fine-tuning results on SR, ACR, NID and SCT with average and variance on 10 runs, which result in small p-value ($<10^{-7}$) and verify significant improvements over the base model. \textbf{Right}: zero-shot evaluation compared to DeBERTa and RoBERTa. 
}
\label{fig:result_var}
\label{fig:zero}
\end{minipage}%
\end{center}
\end{figure}

\paragraph{Setups.}
For continual pre-training, we use RoBERTa-large \citep{Liu2019RoBERTa} as our base model. 
We adopt Adam \citep{KingmaB2014Adam} with a learning rate of 1e-4. 
The maximum training step and the linear warmup step of the learning rate are set to 200K and 5k. Dropout ratio and batch size are 0.1 and 200. Each sample includes a positive example, 5 negatives for correlation-based event ranking and 5 negatives for discourse relation ranking. The weight decay is set to 0.01. The maximum sequence length is 128. Pre-training is conducted on 8$\times$A100 GPUs and takes $\sim 3$ days. For downstream fine-tuning, we run 10 random seeds and keep the fine-tuned model with best dev accuracy for test.
We use Adam optimizer with a learning rate of 1e-5 and a linear warmup step of 1k. The dropout, batch size, weight decay and gradient clipping are 0.1, 32, 0.01 and 1.0, respectively. 

\subsection{Main Evaluation and Quantitative Analysis}

\paragraph{Supervised Fine-tuning. }
We take accuracy as the evaluation metric. Table \ref{tab:sr}, \ref{tab:acr}, \ref{tab:nid} and \ref{tab:sct} list comparison of supervised fine-tuning on task-specific training data for the four datasets. It is shown that EventBERT significantly outperforms RoBERTa-large on all datasets (mean, variance and p-value are in Figure~\ref{fig:result_var}(left)). For example, it outperforms RoBERTa by 2\% on script reasoning, and the improvement on story cloze test is larger (i.e., 4\%). EventBERT achieves state-of-the-art performance on script reasoning, narrative incoherence detection, and story cloze test, and it outperforms all baselines of same model size and without additional data.
EventBERT's superiority demonstrates that EventBERT is a general event-based correlation model, and can be applied to a wide range of downstream tasks.

\begin{figure*}[t]
    \centering
    \includegraphics[width=\textwidth]{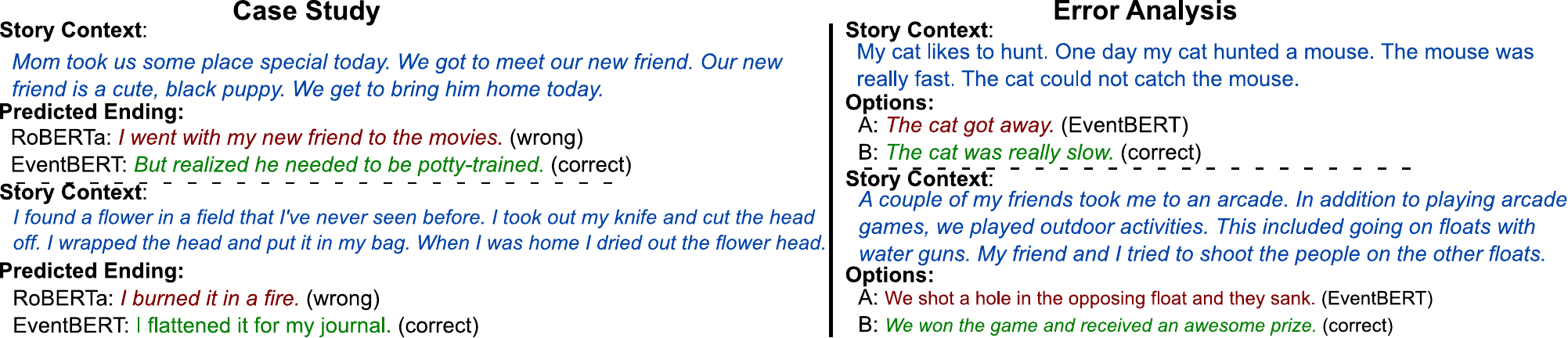}
    \caption{\small Case study (left) and error analysis (right).}\label{fig:case_study}
\end{figure*}

\begin{figure}[t]
\begin{center}
\centering
\begin{minipage}[t]{\linewidth}
\centering
\includegraphics[width=0.52\linewidth]{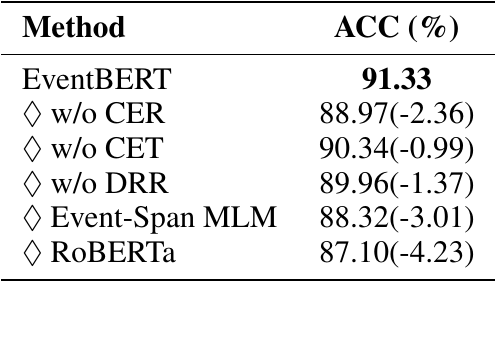}
\includegraphics[width=0.46\linewidth]{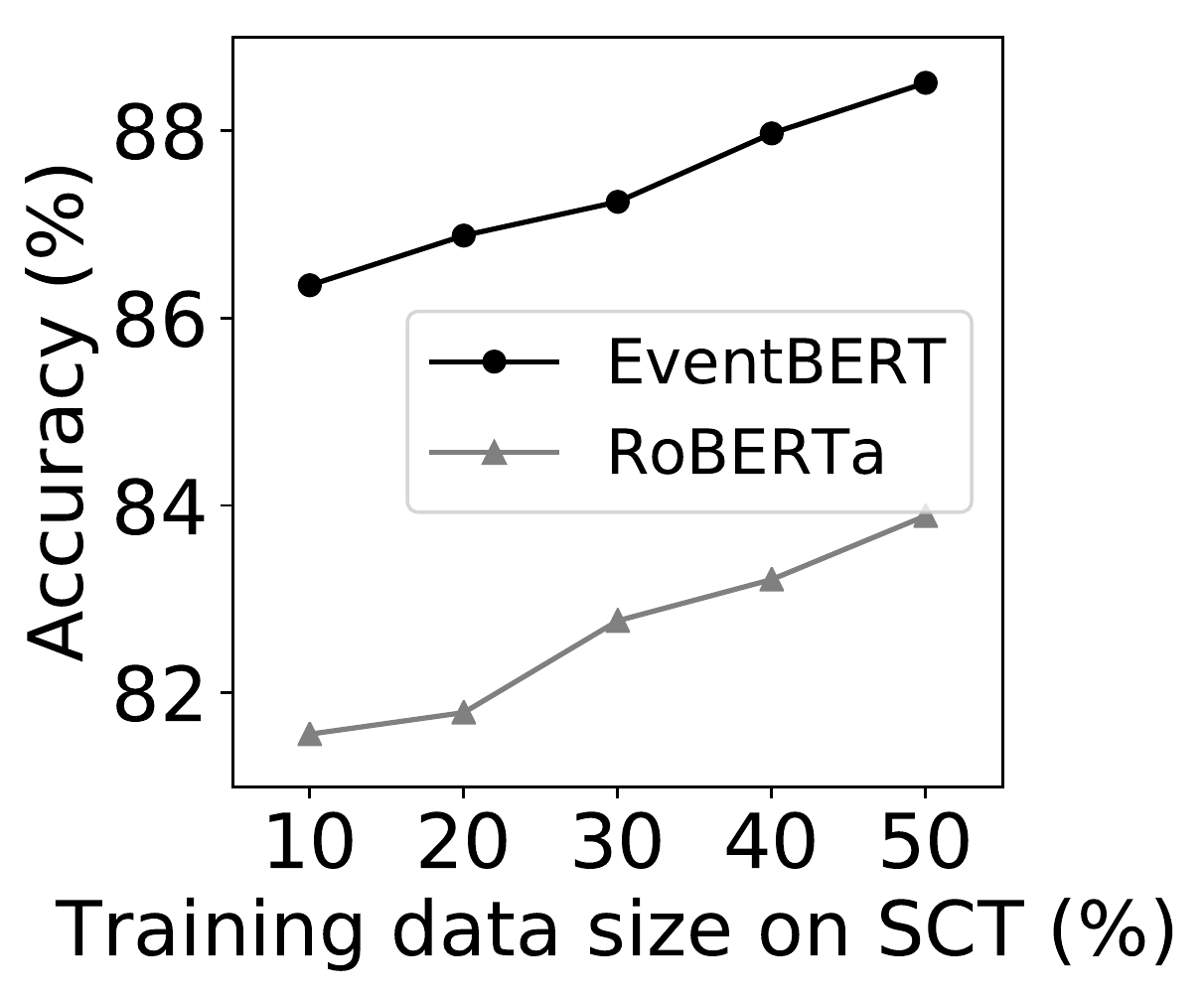}
\caption{\small \textbf{Left}: ablation on story cloze test. \textbf{Right}: impact of training data size.}
\label{fig:few}
\label{fig:abl}
\end{minipage}
\end{center}
\end{figure}

\begin{figure}[t]
\begin{center}
\centering
\begin{minipage}[t]{\linewidth}
\centering
\includegraphics[width=0.52\linewidth]{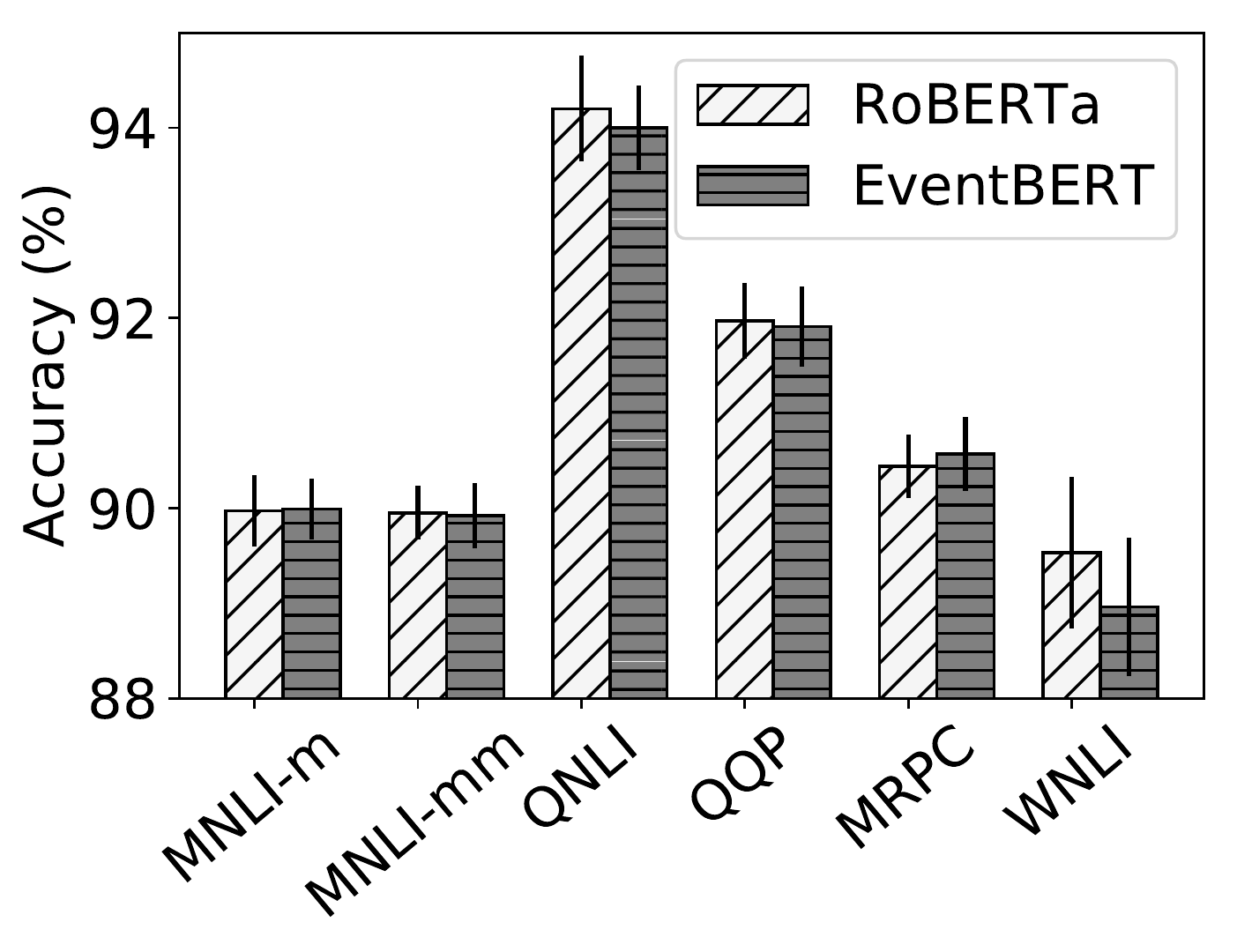}
\includegraphics[width=0.46\linewidth]{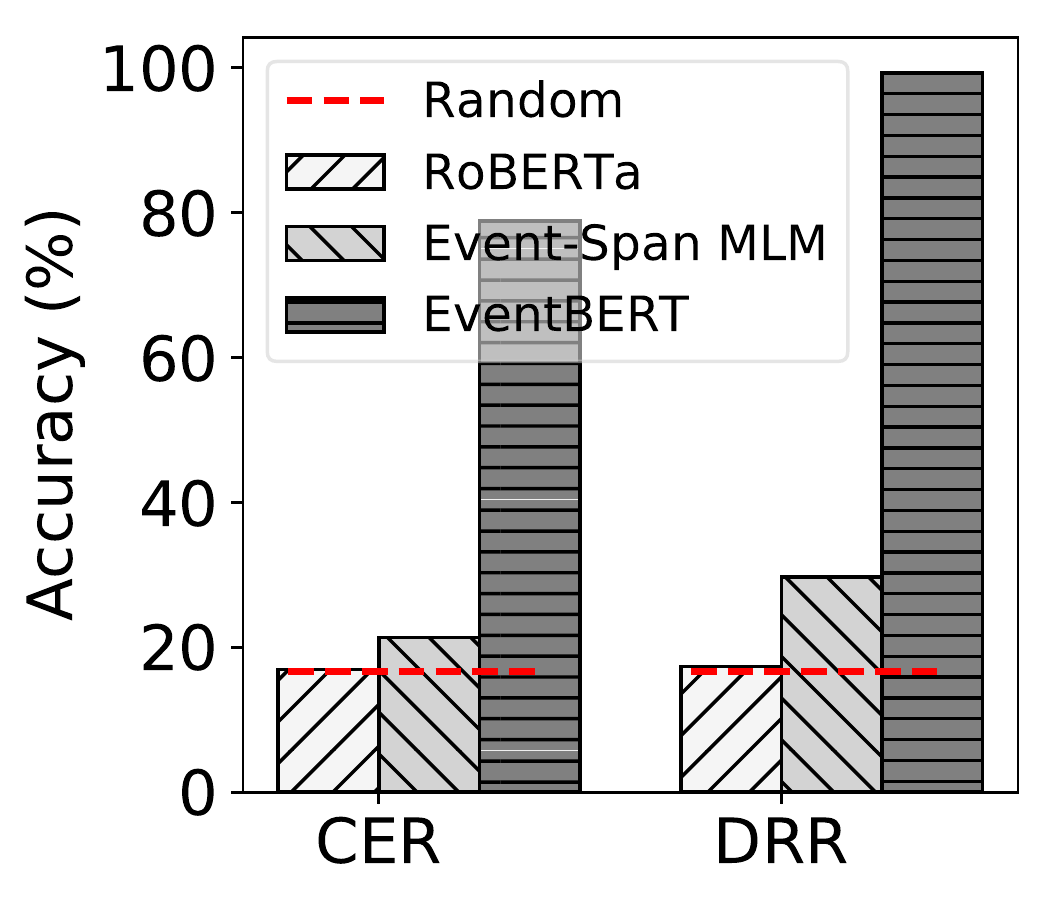}
\caption{\small \textbf{Left}: results on natural language understanding tasks. \textbf{Right}: CER/DRR accuracy on held-out test set (random split of 2\%, about 77k examples) from our pre-training data.}
\label{fig:glue}
\label{fig:data_quality}
\end{minipage}
\end{center}
\end{figure}

\paragraph{Zero-shot Evaluation.}
We apply EventBERT and competitors to downstream tasks without fine-tuning (as in \S\ref{sec:meth-eventbert_eval}). 
Figure~\ref{fig:zero}(right) shows results on script reasoning (SR), abductive commonsense reasoning (ACR), and story cloze test (SCT). RoBERTa/EventBERT is inapplicable to narrative incoherence detection in 0-shot due to an extra MLP needed. 
It is shown RoBERTa performs similarly to random guess, while EventBERT surpasses them by a large margin. RoBERTa only gets 52.6\% accuracy, and EventBERT outperforms it by 23\% absolute value, verifying EventBERT's effectiveness when no task-specific training data is available. 



\paragraph{Ablation Study.}
We conduct an ablation study on story cloze test in Figure \ref{fig:abl}(left) to investigate the effect of each component. We first remove each of the three objectives in \S\ref{sec:meth-eventbert_pretrain} during the pre-training. It is shown, the correlation-based event ranking (CER) objective plays a critical role, and the accuracy drops by 2.36\% w/o it. Besides, ablating contradiction event tagging (CET) and discourse relation ranking (DRR) also lead to 0.99\% and 1.37\% decrease. We also replace the proposed three objectives with existing span-based masked language model for pre-training, which is denoted as ``Event-Span MLM''. Compared to our proposed event correlation based objectives, traditional span-based masked language model cannot fully leverage the event-based corpus, and EventBERT outperforms it by 3\%. 

\paragraph{Impact of Training Data Size.}
Figure \ref{fig:few}(right) plots the accuracy of RoBERTa and EventBERT on story cloze test with various sizes of training data. First, when more training data is used, accuracy of both increases since they can learn more domain-/task-specific knowledge. Second, EventBERT outperforms RoBERTa by a large margin (i.e., 4\%\textasciitilde6\%), and the margin is consistent, which verifies EventBERT can conduct event-based reasoning better with less task-specific data.

\paragraph{Capability Retention for Natural Language Understanding (NLU) Tasks.}
To verify EventBERT is still competitive in NLU after continual pre-training with our objectives, we fine-tuning RoBERTa and EventBERT on some NLU tasks (natural language inference and relatedness), and report average and variance on 5 runs on the dev set in Figure \ref{fig:glue}(left). 

\paragraph{Insight into Built Corpus. } 
To check the quality of our created corpus $\sD$ and verify the incompetence of MLM on event learning, we conduct evaluations on CER and DRR in Figure~\ref{fig:data_quality}(right). We use accuracy (equivalent to Hits@$1$) as the metric and these evaluations follow \S\ref{sec:meth-eventbert_eval}. 
MLM-based RoBERTa (already pre-trained on whole \textsc{BookCorpus})  underperforms on correlation-based masking, even on (almost) word-level discourse relation masking, demonstrating RoBERTa is unlikely to learn event-correlation knowledge, and our built data is non-trivial and challenging to prior pre-trained models. Moreover, even fine-tuning RoBERTa on our corpus via MLM (i.e., Event-Span MLM), the performance is still far from satisfying, verifying MLM is incompetent in correlation learning.

\subsection{Case Study, Error Analysis and Limitation}
\paragraph{Case Study.} Figure~\ref{fig:case_study}(left) shows examples where EventBERT can find the correct ending for a story, while the baseline selects a wrong option. The main reason might be that RoBERTa is more concerned with token-level concurrence while EventBERT takes events as units and focuses on correlation among them. For example, RoBERTa chooses ``\textit{I burned it in a fire}'' for the 2nd case, which might be due to strong correlation between \textit{dry} and \textit{burn}. In contrast, EventBERT understands that ``\textit{I found a flower in a field that I've never seen before}'' and ``\textit{I dried out the flower head}'', and infers that it is more possible that ``\textit{I flattened it for my journal}'.

\paragraph{Error Analysis.} 
We also show some error cases in Figure~\ref{fig:case_study}(right). As we can see that it is difficult to select the correct option for some examples. Take the first case as an example, although according to the context we can infer that it is more likely to choose ``\textit{The cat was really slow}'', the option itself is somehow contradictory to common sense since usually a cat is fast, leading to confusion during inference. We leave such complex situations for future work.

\paragraph{Limitation.} 
First, EventBERT is focused on correlation-based event reasoning, and not general enough to every event correlation reasoning task (e.g., event temporal reasoning as in \citep{Han2020DEER,Lin2020Conditional}).
Second, we evaluate EventBERT on deterministic tasks, e.g., multi-choice and Cloze-type question answering, due to their stable metrics and widely-available baselines. 
Third, we only access limited computation resources and perform continual pre-training from RoBERTa, while previous works \citep{Gu2020PubMedBERT} verify learning with task-specific corpora and objectives from scratch greatly improves.
%

\section{Conclusion} 
We propose to pre-train a general model for event correlation reasoning from unlabeled text. To achieve that, we create a corpus by filtering out paragraphs without strong event correlation and further extracting events for remaining ones. Then we present three correlation-based self-supervised objectives for pre-training. The derived model, EventBERT, outperforms strong baselines on 4 downstream tasks in both zero-shot and supervised fine-tuning settings. 

\bibliographystyle{acl_natbib}
\bibliography{ref}

\appendix

\end{document}

%% file: math_commands.tex
\usepackage{amsmath,amsfonts,bm}

\def\eqref#1{equation~\ref{#1}}

\def\1{\bm{1}}

\def\vh{{\bm{h}}}

\def\vv{{\bm{v}}}

\def\mH{{\bm{H}}}

\DeclareMathAlphabet{\mathsfit}{\encodingdefault}{\sfdefault}{m}{sl}
\SetMathAlphabet{\mathsfit}{bold}{\encodingdefault}{\sfdefault}{bx}{n}

\def\gG{{\mathcal{G}}}

\def\gL{{\mathcal{L}}}

\def\sA{{\mathbb{A}}}

\def\sD{{\mathbb{D}}}
\def\sF{{\mathbb{F}}}

\def\sK{{\mathbb{K}}}
\def\sL{{\mathbb{L}}}
\def\sM{{\mathbb{M}}}
\def\sN{{\mathbb{N}}}

\def\sQ{{\mathbb{Q}}}

\def\sV{{\mathbb{V}}}

\newcommand{\R}{\mathbb{R}}

\newcommand{\softmax}{\mathrm{softmax}}

